\documentclass[twocolumn]{article}
\usepackage[colorlinks,linkcolor=blue]{hyperref}
\usepackage{geometry}
\usepackage{cite}
\usepackage{amsmath,amssymb,amsfonts}
\usepackage{algorithmic}
\usepackage{graphicx}
\usepackage{textcomp}
\usepackage{amssymb}
\usepackage{booktabs}
\usepackage{multirow}
\usepackage{subfigure}
\usepackage{caption}
\usepackage{epstopdf}
\usepackage{subfigure}
\usepackage{changepage}

\def\BibTeX{{\rm B\kern-.05em{\sc i\kern-.025em b}\kern-.08em
    T\kern-.1667em\lower.7ex\hbox{E}\kern-.125emX}}
\begin{document}
\title{3D RoI-aware U-Net for Accurate and Efficient\\ Colorectal Tumor Segmentation}
\author{Yi-Jie Huang$^{1,2}$, Qi Dou$^{3}$, Zi-Xian Wang$^{4}$, Li-Zhi Liu$^{4}$, Ying Jin$^{4}$, Chao-Feng Li$^{4}$,\\
	Lisheng Wang$^{1}$$^{\star}$, Hao Chen$^{2}$$^{\star}$, Rui-Hua Xu$^{4}$$^{\star}$\\
	\small $^{1}$Institute of Image Processing and Pattern Recognition, Department of Automation, Shanghai Jiao Tong University, China\\
	\small $^{2}$Imsight Medical Technology Co. Ltd., China\\
	\small $^{3}$Department of Computer Science and Engineering, The
	Chinese University of Hong Kong, Hong Kong\\
	\small $^{4}$Sun Yat-sen University Cancer Center; State Key Laboratory of Oncology in South China; \\
	\small Collaborative Innovation Center for Cancer Medicine, Guangzhou, China\\
}
\date{}
\maketitle

\begin{abstract}

Segmentation of colorectal cancerous regions from 3D Magnetic Resonance~(MR) images is a crucial procedure for radiotherapy which conventionally requires accurate delineation of tumour boundaries at an expense of labor, time and reproducibility.  
{\color{black}While deep learning based methods serve good baselines in 3D image segmentation tasks, small applicable patch size limits effective receptive field and degrades segmentation performance.} In addition, Regions of interest~(RoIs) localization from large whole volume 3D images serves as a preceding operation that brings about multiple benefits in terms of speed, target completeness, reduction of false positives. Distinct from sliding window or non-joint localization-segmentation based models, we propose a novel multi-task framework referred to as 3D RoI-aware U-Net~(3D RU-Net), for RoI localization and in-region segmentation where the two tasks share one backbone encoder network. With the region proposals from the encoder, we crop multi-level RoI in-region features from the encoder to form a GPU memory-efficient decoder for detail-preserving segmentation {\color{black} and therefore enlarged applicable volume size and effective receptive field}. To effectively train the model, we designed a Dice formulated loss function for the global-to-local multi-task learning procedure. Based on the efficiency gains demonstrated by the proposed method, we went on to ensemble models with different receptive fields to achieve even higher performance costing minor extra computational expensiveness. Extensive experiments were subsequently conducted on 64 cancerous cases with a four-fold cross-validation, and the results showed significant superiority in terms of accuracy and efficiency over conventional state-of-the art frameworks. In conclusion, the proposed method has a huge potential for extension to other 3D object segmentation tasks from medical images due to its inherent generalizability. The code for the proposed method is publicly available.

\end{abstract}

3D CNN, region of interest, multi-task learning, tumor segmentation, colorectal cancer.

\section{Introduction}\label{introduction}
\label{sec:introduction}

\begin{figure}[ht]
	\centering
	\centerline{\includegraphics[width=8.0cm]{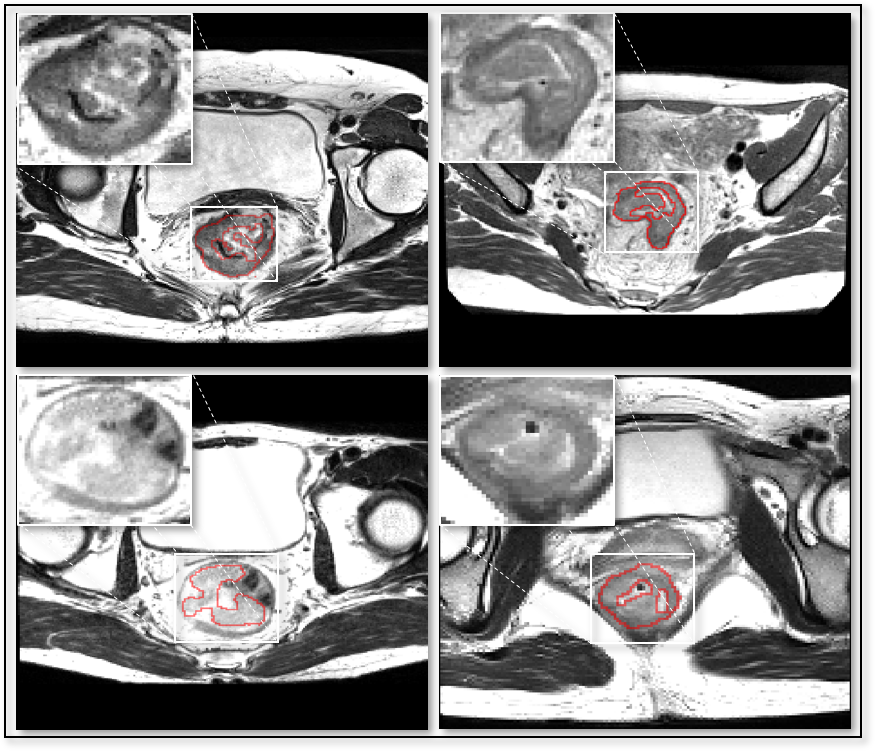}}
	\caption{Typical examples of MR slices with colorectal cancer. The cancer regions are delineated with red lines and zoomed in for clear illustration. Clearly, the target areas cannot be well separated by intensity clipping, shape models or positional priors.}
	\label{fig:Case}
\end{figure}
Colorectal cancer {\color{black}strikes more than 1.4 million people and accounts for 694,000 deaths globally in 2012~\cite{Feb2015World}. It is more common in developed countries, for example, in the USA, colorectal cancer is the second leading cause of cancer-related mortalities~\cite{CAAC:CAAC21387}.} In current clinical routine of radiotherapy, colorectal cancer regions are manually recognized and delineated from volumetric images acquired by magnetic resonance~(MR) imaging for treatment including surgery and radiation therapy. However, this procedure is laborious, time-consuming and observer-dependent, thus suffers from tedious effort and limited reproducibility. Therefore, automatic colorectal tumor detection and segmentation methods are highly demanded to improve the clinical routine. 

Such demand defines a task of automatic detection and segmentation of the targets from whole 3D image volumes. Compared to processing manually selected RoI patches, the superiority of being fully automatic simplifies the workflow, excludes manual intervention and enables fast processing of large amounts of image volumes. Taking initial works based on super-voxel clustering~\cite{mahapatra2013automatic,irving2014automated} one step further, deep learning based methods dominate the state-of-the-art of detection and segmentation field. {\color{black}However, deep learning based methods for this task are challenged by following factors: weak intensity specificity, absence of shape characteristic, lacking positional priors~(as is illustrated in Fig. \ref{fig:Case}), class imbalance and long processing time of existing methodologies under inferior GPU or CPU-only deployment environments. 

Apart from aforementioned challenges, a vital 3D image specific problem is not fully tackled by the community. Among existing methods for fully automatic image segmentation~\cite{badrinarayanan2017segnet,Chen20163D,ronneberger2015u, cciccek20163d,milletari2016v,yu2017volumetric,Chen2017VoxResNet,dou20173d}, though a plausible performance can be achieved by utilizing multi-level features~($e.g.$ use skip connections) to gather fine grained details that are lost in the down-sampling process, the merit of maintaining a global understanding represented by deep features with large receptive field is not fully enjoyed due to patch size limitation of GPU memory. As is supported by many researches for 2D image processing, $e.g.$, dilated convolutions~\cite{yu2015multi} and pyramid pooling schemes~\cite{zhao2017pyramid}, enlarging receptive fields enables wide-range context utilization and makes further performance breakthroughs. In medical applications, global understanding is even more important since that the targets and the background are highly correlated.}




Generally, existing methods for lesion detection and segmentation from 3D images can be divided into part based models and non-joint localization-segmentation based models.

Initially, as naive practices, part based FCNs learn from local parts of 2D slices~\cite{ronneberger2015u,wang2018deep,chen2016combining}, 2.5D slices~\cite{Roth2015DeepOrgan,Roth2014A} or small 3D patches~\cite{yu2017volumetric,Nie20183} and perform (often overlapped) part-sliding for whole volume inference, which is slow and prone to false positives and target incompleteness related failures. {\color{black}More importantly, part based methods suffer from limited effective receptive fields. V-Net~\cite{milletari2016v}, for example, claimed $551\times 551 \times 551$ designed receptive field but used $64\times 128 \times 128$ patch sliding scheme, making the large designed receptive field not fully effective. To enlarge the effective receptive field under current part based frameworks, Crossbar-Net~\cite{yu2018crossbar} proposed to train segmentation networks using non-squared patches with different aspect ratios to add more global contexts to local details.}

More recently, trends highlight potential accuracy and speed benefits of adding RoI localization modules prior to FCNs. As a common practice, the RoI localization modules are individually designed as a standalone part of a pipeline. Conventionally, RoIs are localized using prior knowledge such as multi-atlas registration, which is often used to localize normal organs~\cite{Rohlfing2004Performance,Klein2008Automatic}. {\color{black}Apart from their inappropriateness for lesion localization, they are relatively slow. As is reported in \cite{murphy2014fast}, registration takes at least 20 seconds per patient using GPUs and typically tens of minutes per patient using CPUs.} Learning based RoI localization decouples RoI localization from prior knowledge~\cite{Hariharan2015Hypercolumns,Dai2015Convolutional,Pinheiro2015Learning,li2017h,Liao2017Estimation}. Some of the related practices\cite{Hariharan2014Simultaneous,Hariharan2015Hypercolumns} extract region proposals using external modules such as Selective Search~\cite{Uijlings2013Selective} or Multiscale Combinatorial Grouping~(MCG)~\cite{arbelaez2014multiscale}, {\color{black}which are also well-known speed bottlenecks as is pointed out in ~\cite{Ren2015Faster} and replacing them with RPN accelerated a network from 0.5 fps to 5 fps.} Later works adopt light CNN models such as 2D CNNs for RoI localization and 3D FCNs for in-region segmentation~\cite{li2017h,tang2018segmentation,Balagopal2018Fully}. Compared to part based methods, these works tackle the tasks in more graceful manners. {\color{black}Still, using a standalone FCN for RoI segmentation requires repeated extraction of low-level features without possible feature sharing, yet feature sharing is reported in \cite{Girshick2015Fast} to produce 213X acceleration for object detection, given large numbers of target candidates. Nonetheless, using a patch-based FCN for RoI segmentation leaves the problem of limited effective receptive fields unsolved.}

As a promising development, joint RoI localization-segmentation models such as Multi-task Network Cascades~(MNC)~\cite{Dai2016Instance} and Mask R-CNN~\cite{He2017Mask} further eliminate redundant feature extraction and achieve better speed and accuracy by sharing a backbone network across the sub-nets for region proposal, region classification and in-region segmentation. Mask R-CNN employs Feature Pyramid Network~(FPN)~\cite{lin2017feature}, which is encoder-decoder-skip connection formulated, and used scale-specific feature maps for better segmentation details. {\color{black}An apparent drawback of Mask R-CNN is that using scale-specific feature maps and RoIAlign's bin-fitting scheme for segmentation are still detail-lossing, though better than a non-FPN version; To tackle this issue, PA-Net\cite{Shu2018Path} added another bottom-up path for better segmentation detail, which is even more costly for a 3D application. Another drawback of direct extending it to 3D lies on the need of forming anchor boxes defied by additional aspect ratios along the Z axis. Fitting a small amount of 3D objects to more anchor boxes is prone to bad-shaped bounding box prediction.}



Apart from the way whole volume predictions are generated, recent works propose some strategies to further boost the performance of volumetric tasks. Firstly, V-Net~\cite{milletari2016v} adopts parameter-free Dice coefficient~\cite{Dice1945Measures} loss to harness the class-imbalance issue. Secondly, inspired by the success of multi-task learning\cite{Zhang2014Facial,chen2017ultrasound}, Deep Contour-aware Networks~(DCAN)~\cite{chen2016dcan} and Boundary-aware FCN~\cite{Shen2017Boundary} employ contour-aware loss functions for better discrimination between boundaries and the background. In addition, Multilevel Contextual 3D CNNs~\cite{Qi2017Multilevel}, DeepMedic~\cite{Kamnitsas2016Efficient}, Orchestral Fully Convolutional Networks~(OFCNs)~\cite{xu2018orchestral} and Hybrid Loss guided Fully Convolutional Networks~(HL-FCNs)~\cite{Huang2018HL} adopt model ensemble for better robustness. 
\begin{figure*}[ht]
	\centering
	\centerline{\includegraphics[width=17.6cm]{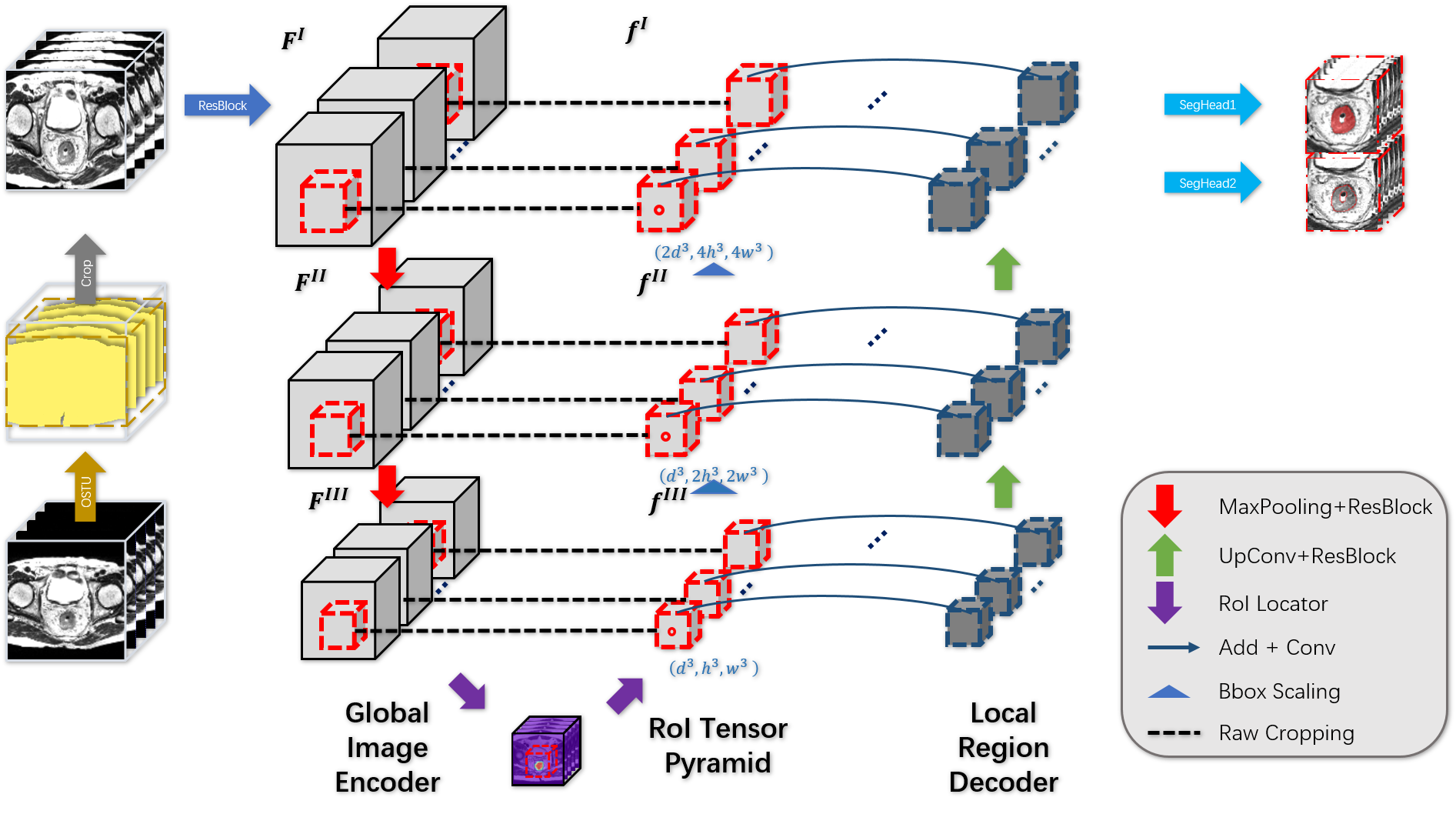}}
	\caption{The illustration of 3D RU-Net. The network consists of the Global Image Encoder, the RoI Tensor Pyramid and the Local Region Decoder. A bounding box is predicted using feature maps $\textbf{F}^{III}$ and are extended as a Bounding Box Pyramid, then the corresponding RoI Tensor Pyramid $(f^I,f^{II},f^{III})$ is extracted from $(F^{I},F^{II},F^{III})$ and memory-efficient multilevel feature fusion for in-region segmention is performed in the decoder stage.}
	\label{fig:R-UNet}
\end{figure*}

A part based initial work to automatically segment colorectal cancer regions was published in ISBI~\cite{Huang2018HL}. {\color{black}As a step further, in this paper, we propose a novel joint RoI localization-segmentation framework named as 3D RoI-aware U-Net~(3D RU-Net) to enjoy the benefits of fast RoI localization, target completeness and large effective receptive field of joint detection-segmentation frameworks while maintaining the easy-to-train and detail-preserving merits of popular end-to-end and volume-to-volume segmentation methods.} To effectively train the model, we design a hybrid loss function to help the network both handle small objects in big volumes and focus on accurately recognizing ambient borders in local RoIs, and additionally adopt low-cost {\color{black}multi-receptive field} ensemble strategy for better robustness. Experiments conducted on 64 acquired scans demonstrated the efficacy of our method and ablation studies validate the contribution gain of each component from our framework.

Our main contributions are summarized as follows:

\begin{enumerate}
	\item We propose a 3D joint RoI localization-segmentation framework with a shared Global Image Encoder for global-understanding based RoI localization, and a Local Region Decoder working on pyramid-designed in-region features for RoI segmentation. This design enables fast and memory efficient detail-preserving whole volume segmentation {\color{black}with full use of large receptive fields} compared to its competing counterparts.
	\item Considering automatic class rebalancing and better boundary discrimination, we propose a Dice formulated global-to-local multi-task hybrid loss~(MHL) function to further improve the accuracy. Additionally, the accelerated framework encourages us to employ a multiple receptive field model ensemble strategy to suppress the false positives and refine the boundary details at an acceptable speed cost.
	\item Extensive experiments on the acquired dataset proved the efficacy of our proposed framework. Furthermore, our method is inherently general and can be applied in other similar applications. 
\end{enumerate}

The remainder of this paper is organized as follows. We describe
our method in Section II and report the experimental
results in Section III. Section IV further discusses some insights as well as issues of the proposed method. The conclusions are drawn in Section V.

\section{Methodology}

In this section, to address slow prediction and {\color{black}limited effective receptive field} issues of non-joint models along with detail-lossing and bad bounding box issues of joint models discussed in Section~\ref{introduction}, we propose a framework to effectively localize and segment colorectal tumors from whole volume 3D images.

\subsection{Construction of 3D RU-Net}

The proposed 3D RU-Net architecture is illustrated in Fig. \ref{fig:R-UNet}. We input whole image volumes to Global Image Encoder for multi-level feature encoding, {\color{black} employ an encoder-only RoI locator for RoI localization}, crop in-region feature tensors from multi-scale feature maps {\color{black} using RoI Pyramid Layer}, and design a Local Region Decoder sub-network to perform multi-level feature fusion for high-resolution cancerous tissue segmentation.

\subsubsection{Global Image Encoder} \label{Design}Due to limited GPU memory of commonly used devices and dramatically increased parameters of 3D convolution kernels, it's essential to carefully design the 3D backbone feature extractor to avoid GPU memory overflow and overfitting. 

Instead of constructing a complete 3D version of encoder-decoder architecture like 3D FPN, or directly extending popular backbones~\cite{simonyan2014very,He2016Deep,huang2017densely} to 3D, a compact encoder-only network named the Global Image Encoder is constructed to process whole volume images rather than dealing with context-limited small parts as common practices do. {\color{black} Specifically, the encoder employs a stack of ResBlocks~\cite{He2016Deep} and MaxPooling layers to encode whole volume images. Each Residual Block has three convolutional layers, three Instance Normalization Layers~\cite{Ulyanov2016Instance}, three ReLU layers and a Skip Connection for better gradient flowing. The Instance Normalization Layer is used for better robustness given $batchsize=1$ in 3D segmentation tasks.}

{\color{black}\subsubsection{RoI Locator} The RoI Locator is a template where any method that employs encoder-only backbones for target detection can be employed. Due to aspect ratio diversity of number-limited training samples, learning accurate bounding box regression can be difficult. For this specific 3D semantic segmentation task, we recommend taking full advantage of available voxel-level masks as is discussed below for simplicity and more robust bounding box prediction.

Specifically, we avoid degrading voxel-wise labels to object-wise labels to learn anchor fitting. Instead, the locator is designed as a module taking feature map $F^{III}$ as input, consisting of a convolutional layer with kernel size $1$ and $Sigmoid$ activation function. This module is trained to predict down-sampled segmentation masks from global images. To tackle the extremely imbalanced foreground-to-background ratio, instead of partial sampling, i.e. sampling a fixed proportion of foreground and background or employing OHEM\cite{shrivastava2016training}, the locator is trained towards Dice loss, which will be introduced in subsection \ref{Dice}. Then we perform a fast 3D connectivity analysis to compute desired bounding boxes  formulated as  $Bbox^{III}=(z^3,y^3,x^3,d^3,h^3,w^3)$ where $(z^3,y^3,x^3)$ denotes the starting coordinates and $(d^3,h^3,w^3)$ denotes depth, height and width of $Bbox^{III}$ in feature map $F^{III}$.}

\begin{figure}[ht]
	\centering
	\centerline{\includegraphics[width=8.5cm]{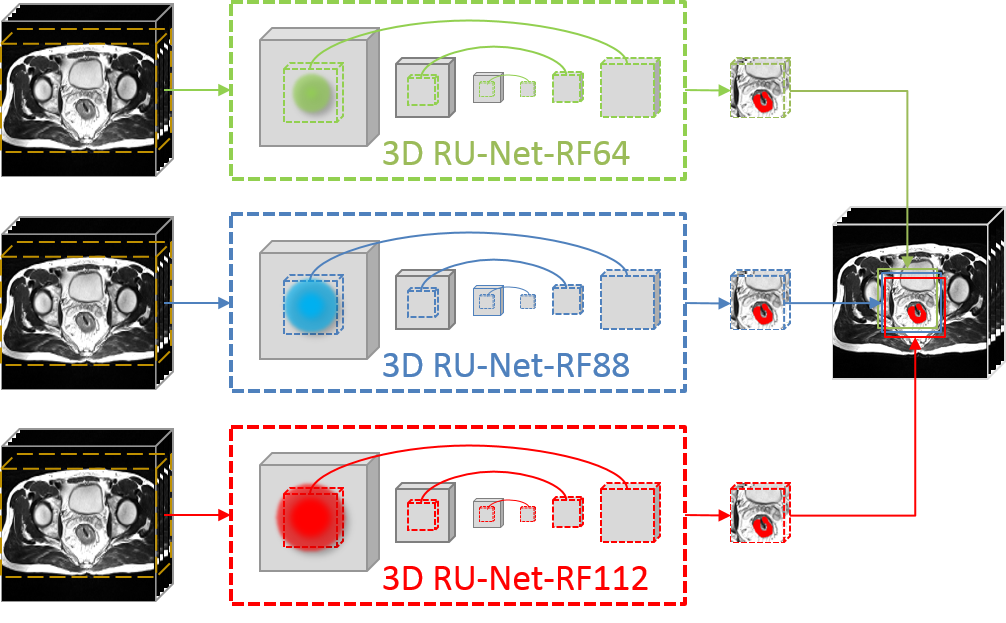}}
	\caption{3D RU-Net-RF64, 3D RU-Net-RF88 and 3D RU-Net-RF112 are of different dilation rates. The green, blue and red spheres of different sizes indicate receptive fields of $26\times 64 \times 64$, $26\times 88 \times 88$ and $26\times 112 \times 112$, respectively. In the output end, their predictions are averaged.}
	\label{fig:Ensemble}
\end{figure}

{\color{black}\subsubsection{RoI Pyramid Layer} 
As is illustrated in Fig. \ref{fig:R-UNet}, we propose a novel layer named RoI Pyramid Layer. Instead of bin-fitting the RoI tensor cropped from a manually selected single-scale feature map, in this paper, we propose to extract a group of raw multi-level feature tensors from each feature scale named as RoI Tensor Pyramid for full utilization of multi-level features and better mask details.

To extract an RoI Tensor Pyramid for a detected target, we first construct a Bounding Box Pyramid $(Bbox^I, Bbox^{II},Bbox^{III})$ from a given bounding box $Bbox^{III}=(z^3,y^3,x^3,d^3,h^3,w^3)$. Specifically, Bounding Box Pyramid is computed iteratively following Bbox Scaling criterion listed below:
\begin{equation}
\begin{aligned}
Bbox^{i-1}=&(z^i\times s^i_z, y^i\times s^i_y,x^i\times s^i_x,\\
&d^i\times s^i_z, h^i\times p^i_y,w^i\times s^i_x)
\end{aligned}
\end{equation}
where $(s^i_z,s^i_y,s^i_x)$ denotes the stride configuration of $MaxPooling^i$ layer. Given the Bounding Box Pyramid $(Bbox^I, Bbox^{II},Bbox^{III})$, we crop raw RoI Tensor Pyramid $(f^I,f^{II},f^{III})$ from whole volume feature maps $F^I, F^{II}$ and $F^{III}$ without applying any bin-fitting operation and form an RoI Tensor Pyramid for posterior Local Region Decoder branch.
}
\subsubsection{Local Region Decoder}

Given a RoI Tensor Pyramid, we construct a sub-network for in-region segmentation named as Local Region Decoder by applying successful multilevel feature fusion mechanism. The construction of the decoder is more or less symmetrical to the encoder part with skip connections to fuse feature maps of corresponding scales, while the beneficial difference lies on much smaller sizes of the decoder branch's feature tensors. Since no shape distortion or scale normalization is included in the RoI Pyramid Layer, this module restores the spatial dimension of the RoI region without lossing details. The same set of decoder weights is used to iteratively process different RoIs if multiple RoIs are localized.

\subsection{Dice-based Multi-task Hybrid Loss Function}\label{Dice}

In multi-task learning practices, each task faces different challenges. In our case, the Global Image Encoder mainly suffers from class imbalance issue, while the Local Region Decoder has to focus on the exact boundaries of the target regions. Thus we propose a Dice-based multi-task loss~(MHL) function to effectively learn these tasks.

\begin{figure}[ht]
	\centering
	\centerline{\includegraphics[width=8.5cm]{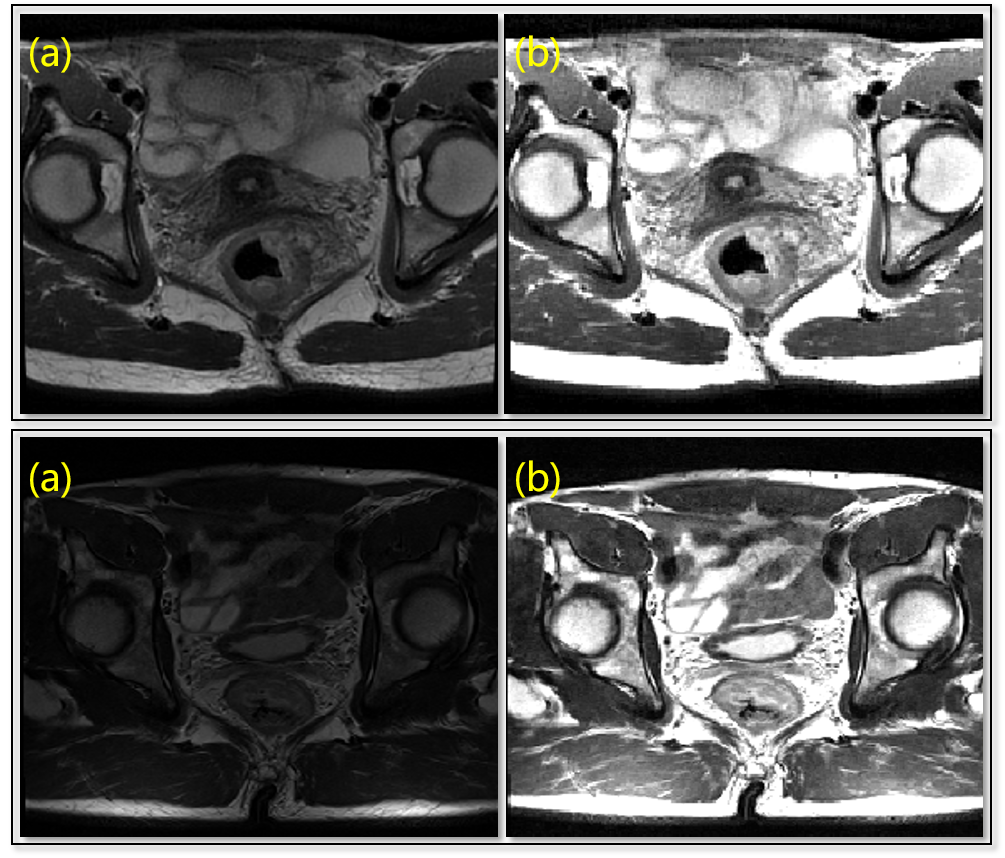}}
	\caption{Examples of: (a) Original Images (b) Normalized Images. The intensity of homogeneous tissues from images acquired under different imaging configurations are normalized to identical ranges.}
	\label{fig:Preprocessing}
\end{figure}
\subsubsection{Dice Loss Formulation} 
Inspired by the success of \cite{milletari2016v}, we apply Dice loss function to formulate the optimization objective, since it serves as an effective hyper-parameter free class balancer to help the network learn objects of small size and weak saliency. The Dice loss is defined as:
\begin{equation}
L_d(P,G)=1-2\times\frac{\sum_{i=1}^{N}{p_{i}g_{i}}+\epsilon}{\sum_{i=1}^{N}{p_{i}}+\sum_{i=1}^{N}{g_{i}}+\epsilon }
\end{equation}
where the sums are computed over the $N$ voxels of the predicted volume $p_{i} \in P$ and the ground truth volume $g_{i} \in G$. $\epsilon$ is a {\color{black}minimal} smoothness term that avoids {\color{black}division} by 0 {\color{black}and is set as $10^{-4}$}. In the optimization stage, the Dice loss is minimized by gradient descen{\color{black}t} using the following derivate:
\begin{equation}
\frac{\partial L_{d}(P,G)}{\partial p_{k}}=  -2\times \frac{\sum_{i=1}^{N}p_{i} g_{i} -g_{k}\sum_{i=1}^{N}(p_{i}+g_{i})}{[\sum_{i=1}^{N}(p_{i}+g_{i})]^{2}}
\end{equation}

\subsubsection{Dice Loss for Global Localization}

To tackle the class imbalance issue of the global image RoI localization task, we employ the aforementioned Dice loss:
\begin{equation}
L_{global}=L_d(P_{global},G_{global})
\end{equation}
where $P_{global}$ and $G_{global}$ denotes predictions of the localization top and down-sampled annotations.
\begin{table*}	
	
	\centering
	\resizebox{0.95\textwidth}{!}{
		\begin{tabular}{|c|c|c|c|c|c|c|c}
			\toprule[1pt]
			Part Name & {\color{black}Input Layer} & {\color{black}Module Name} & Kernel & Out Channels & {\color{black}Receptive Field 1}& {\color{black}Receptive Field 2}& {\color{black} Receptive Field 3} \\  
			\midrule
			\multirow{6}{1.5cm}{{\color{black}Encoder}} & Image & ResBlock1 & $1\times3\times3$ & 48 & $1\times 7\times 7$&$1\times 7\times 7$&$1\times 7\times 7$\\
			&ResBlock1   & MaxPooling1 &$1\times1/2\times1/2$ & 48 & -&- &- \\
			&MaxPooling1 & ResBlock2 &  $3\times3\times3$ & 96 & $7\times 20\times 20$&$7\times 20\times 20$&$7\times 34\times 34$\\
			&ResBlock2   & MaxPooling2 &  $1/2\times1/2\times1/2$ & 96& - & - & - \\
			&MaxPooling2 & ResBlock3 &  $3\times3\times3$ &192 &$20\times 46\times 46$ & $20\times 70\times 70$ & $20\times 46\times 46$\\
			& ResBlock3  & Locator~(sigmoid) &$1\times1\times1$& 1 & $20\times 46\times 46$& $20\times 70\times 70$ & $20\times 82\times 82$\\
			\midrule
			\multirow{3}{1.5cm}{{\color{black}RoI Pyramid Layer}} & {\color{black}Locator},ResBlock1  & {\color{black}RoI Tensor I} & - &48&$1\times 7\times 7$ & $1\times 7\times 7$&$1\times 7\times 7$\\
			& {\color{black}Locator},ResBlock2 &{\color{black}RoI Tensor II} & - &96&$7\times 20\times 20$&$7\times 20\times 20$&$7\times 34\times 34$\\
			& {\color{black}Locator},ResBlock3 &{\color{black}RoI Tensor III} & - &192&$20\times 46\times 46$&$20\times 70\times 70$&$20\times 82\times 82$\\
			\midrule
			\multirow{8}{1.5cm}{Decoder}& {\color{black}RoI Tensor III} & UpConv1 & $2\times2\times2$ &96& - & - & -\\		
			& {\color{black}RoI Tensor II},UpConv1 & Add1 & - &96& - & - & - \\
			& Add1 &ResBlock4 & $3\times3\times3$ &96&$26\times 58\times 58$&$26\times 82\times 82$&$26\times 106\times 106$\\
			& ResBlock4 &UpConv2 & $1\times2\times2$ &48&-&-&-\\
			& {\color{black}RoI Tensor I},UpConv & Add2 2& - &48&-&-&-\\
			& Add2 &ResBlock5 & $1\times3\times3$ &48&$26\times 64\times 64$&$26\times 88\times 88$&$26\times 112\times 112$ \\
			& ResBlock5 &SegHead1~(sigmoid) & $1\times1\times1$&1&$26\times 64\times 64$&$26\times 88\times 88$& $26\times 112\times 112$\\
			& ResBlock5 &SegHead2~(sigmoid) & $1\times1\times1$ & 1 & $26\times 64\times 64$&$26\times 88\times 88$ & $26\times 112\times 112$ \\	
			\bottomrule[1pt]
		\end{tabular}
	}
		
	\caption{Parameters and connectivity of the network.}
	\label{tab:Architecture}
\end{table*}


\subsubsection{Dice-based Contour-aware Loss for Local Segmentation} 

Compared to the localization task, the in segmentation branch needs multiple constraints to acquire better boundary-sensitive segmentation results. In semantic segmentation practices, the ambiguous borders are the most difficult to learn but learned with insufficient attention. Borrowing the insight of previous exploration of adding an auxiliary contour-aware side task\cite{chen2016dcan}, we further formulate the side task using Dice loss to help it tackle the extreme sparsity of contour labels in 3D space. Practically we add an extra {\color{black}$1\times 1 \times 1$ convolutional layer activated by $Sigmoid$ function} at the output terminal of the segmentation branch to predict the contour voxels, trained in parallel with the region segmentation task. Taking the side task into account, the loss function of the segmentation branch {\color{black} $L_{local}$} is denoted as following by summarizing the weighted losses:

\begin{equation}
\begin{aligned}
L_{local}=&L_{d}(P_{region},G_{region})+\\
&\lambda_{c}L_{d}(P_{contour},G_{contour})
\end{aligned}
\end{equation}
where {\color{black}$\lambda_{c}=0.5$}, denoting the auxiliary task weight to ensure that the region segmentation task dominates while other tasks take effects.

Finally, the overall loss function is:
\begin{equation}
L=L_{global}+L_{local}+\beta\left \|W\right \|^2_2
\end{equation}
where $\beta=10^{-4}$ denotes the balance of weight decay term and $W$ denotes the parameters of the whole network.
 
%
		
		

\subsection{{\color{black}Multiple Receptive Field Model Ensemble}}\label{ensemble}

Due to the limited accuracy of single models, ensemble of multiple models is considered as an effective practice to perform robust inference, and is widely employed in practical cases, at a cost of computational expensiveness.

{\color{black}Encouraged by the dramatically accelerated framework, in this paper, we propose to employ multiple receptive field model ensemble strategy by fusing models of identical structure but with different receptive field settings. This is a generalization to the multi-resolution strategy proposed in \cite{Huang2018HL} that applies identical receptive field to images with different spatial resolutions, which is actually formulating different spatial receptive fields. Such generalization gets rid of detail-losing down-sampling and allows each model contribute to boundary details equally.

In detail, as is illustrated in TABLE \ref{tab:Architecture}, we first construct an original 3D R-U-Net of receptive field $26\times 64 \times 64$, named 3D~RU-Net-RF64. Next, we tune the dilation rate of $ResBlock3$ as 2, enlarging the receptive field to $26\times 88 \times 88$ and formulate 3D~RU-Net-RF88; We further tune the dilation rates of $ResBlock2$, $ResBlock3$ and $ResBlock4$ as 2 and construct a 3D~R-U-Net of receptive field $26\times 112 \times 112$ named 3D~RU-Net-RF112.}

In the inference stage, as is shown in Fig. \ref{fig:Ensemble}, three networks' outputs are averaged to generate the final prediction. {\color{black}Major voting produces similar scores and is therefore not discussed.}

\section{Experiments}
\label{sec:guidelines}

\subsection{Dataset and Preprocessing}
\subsubsection{Dataset} The dataset contains a total of 64 MR images of the pelvic cavity of T2 modality whose ZYX spacings range from ${{3.6}\times{0.31}\times{0.31}}$ mm to ${{4.0}\times{1.0}\times{1.0}}$ mm. Target areas were labeled voxel-wisely by experienced radiologists, and contour labels were automatically generated from the region labels {\color{black}of one-voxel thickness} using erosion and subtraction operations. {\color{black} An 3D image has mostly one and up to two RoIs containing cancerous tissues.}

\subsubsection{Preprocessing} {\color{black}Different spacing rates are normalized to ${{4.0}\times{1.0}\times{1.0}}$ as the HighRes set. Some part-based methods listed in TABLE \ref{tab:compare} employ down-sampled image sets, namely LowRes set of ${{4.0}\times{2.0}\times{2.0}}$ mm spacing and MidRes set of ${{4.0}\times{1.5}\times{1.5}}$ spacing.} To normalize the intensities of input images acquired under different imaging configurations and field of views, we perform in-body intensity normalization to exclude the affect of inconsistent body-to-background ratios. By OTSU\cite{Otsu1979A} thresholding, connectivity analysis and closing operation, body masks are extracted as foreground and other voxels are set as background. The mean intensity and standard deviation are computed within the body mask according to following formulas:

\begin{equation}
Mean(X)=\frac{1}{N_{mask}}\sum_{i\in mask}{x_{i}}
\end{equation}
\begin{equation}
std(X)=\sqrt{\frac{1}{N_{mask}}\sum_{i\in mask}{(x_{i}-Mean(X))^2}}
\end{equation}
where $x_i \in X$ denotes the intensity of a voxel and $N_{mask}$ denotes the count of mask voxels. Then the image is normalized according to {\color{black}standard} normalization criterion.

A few examples of the comparison between original images and intensity-normalized images are illustrated in Fig. \ref{fig:Preprocessing}.
 
Before feeding the images to the network, we crop the input images according to minimum bounding boxes of the body masks to further reduce the GPU memory footprint. Additionally, in the training stage, we performed on-the-fly data augmentation when feeding training samples. Applied random operations include {\color{black}0.9X to 1.1X scaling, flipping w.r.t. the X axis, 0.9X to 1.1X intensity jittering, and RoI translation that shifts the RoI center by -50\% to 50\% width long each axis}.

\subsection{Implementation Details}
Our implementation is publicly available at \url{https://github.com/huangyjhust/3D-RU-Net}.

\subsubsection{Hyper-Parameters} The network's detailed connectivity and kernel configuration are illustrated in Table \ref{tab:Architecture}. Specifically, to fit the anisotropic spacing of the acquired dataset which has larger spacing along Z axis, flat kernels of $1\times3\times3$, pooling rate of $1\times1/2\times1/2$ and up-sampling rate of $1\times2\times2$ are employed by the input and output blocks, $i.e.$ ResBlock1, MaxPooling1, UpConv2, ResBlock5. {\color{black}Initial experiments demonstrate that adding MaxPoolings, ResBlocks or channels does not improve the performance, hence we tune receptive field setting by applying dilated convolution rather than adding layers.} 

\subsubsection{Training Process} The backbone network were initialized using criterion proposed in~\cite{He2015Delving}, then pre-trained using our previous work's patch-wise HL-FCN\cite{Huang2018HL}. We used Adam\cite{Kingma2014Adam} optimizer at a learning rate of $10^{-4}$. The weights of convolution kernels were penalized with $10^{-4}$ L2 norm for better generalization capability. Then, we first train the RoI locator until evaluation loss no longer decrease, then jointly train the RoI locator and the segmentation branch. In each joint training iteration, we accumulate the losses of the RoI Locator, SegHead1 and SegHead2.

\subsection{Evaluation Metrics}

\subsubsection{Dice Similarity Coefficient~(DSC)} The Dice similarity coefficient~(DSC) measures a general overlap rate that equally assigns significance to recall rate and false positive rate. DSC is denoted as:
\begin{equation}
{\color{black} DSC(P,G)=\frac{2|P \cap G|}{\left |P\right |+\left | G \right |}}
\end{equation}
where the metric is scored in [0,1]. Better prediction generates a score closer to 1.0. Since this network is trained towards this metric, DSC is not enough to evaluate the performance.
\subsubsection{Voxel-wise Recall Rate} We also employ voxel-wise recall rate to evaluate the recall capability of different methods.
\begin{equation}
{\color{black}Recall=\frac{|P\cap G|}{|G|}}
\end{equation}

\subsubsection{Average Symmetric Surface Distance~(ASD)} We define the shortest distance of an arbitrary voxel of one volume's surface to another volume's surface as:
\begin{equation}
d(a_k,B)=\min_{b_i \in {\color{black}S(B)}, a_k \in {\color{black}S(A)}}{\left \| a_k-b_i \right \|}
\end{equation}
{\color{black} where $a_k$ denotes $k$th voxel from extracted surface $S(A)$ of volume $A$, $b_i$ denotes $i$th voxel from extracted surface $S(B)$ of volume $B$, and $\left \|.\right \|$ denotes Euclidean distance.} Then the evaluation value is defined as:
\begin{equation}
ASD=\frac{\sum_{p_k\in {\color{black}S(P)}}{d(p_k,G)}+\sum_{g_k \in {\color{black}S(G)}}{d(g_k,P)}}{\left|S(P)\right|+\left|S(G)\right|}
\end{equation} 
{\color{black}where $\left|S(P)\right|$ and $\left|S(G)\right|$ denote the number of surface voxels.}

Specifically, this metric is sensitive to failures such as debris outliers predicted far away from the colon region or complete failure to recall an object. The long distance makes up for the small size of the debris and produce large error penalty. If a failure segmentation has 0 recall rate, its surface distance is set as 50 mm, which is big enough to be a strong penalty.

\subsubsection{Average Inference Time} We include average inference time to evaluate speed in the inference stage. Since this metric is decided by the size of the input volume, the standard deviation is not evaluated. The tested methods are all performed on a workstation platform with 2x Xeon E5 CPU~(8C16T) @ 2.4 Ghz, 128GB RAM and an NVIDIA Titan Xp GPU with 12GB GPU memory. The code is implemented with {\color{black}PyTorch and the inference speed is evaluated under volatile mode}.

\subsubsection{Typical GPU Memory Footprint} By analyzing this metric, we describe the GPU memory efficiency of the proposed methods by tracking the total GPU memory footprint given an input volume of typical size $40\times180\times320$ voxels.  

\begin{figure*}
	\centering	
	\includegraphics[width=\textwidth]{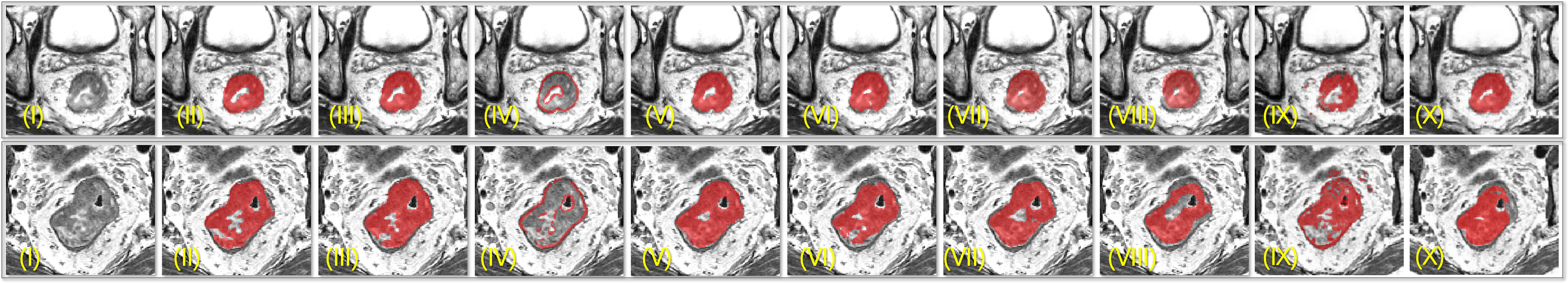}
	\caption{{\color{black} (I) Cancerous region, (II) Expert delineation, (III) Proposed method(predicted regions), (IV) Proposed method (predicted contours) (V) 3D U-Net+DL\cite{milletari2016v} (Ensemble) (VI) 3D U-Net\cite{cciccek20163d} (VII) 3D FCN+3D U-Net (VIII) 3D Mask R-CNN\cite{He2017Mask}(IX) Super-Voxel clustering\cite{irving2014automated}(X) 2D kU-Net+LSTM\cite{chen2016combining}}}
	\label{fig:Results1}
\end{figure*}
\begin{figure*}
    \centering
	\includegraphics[width=\textwidth]{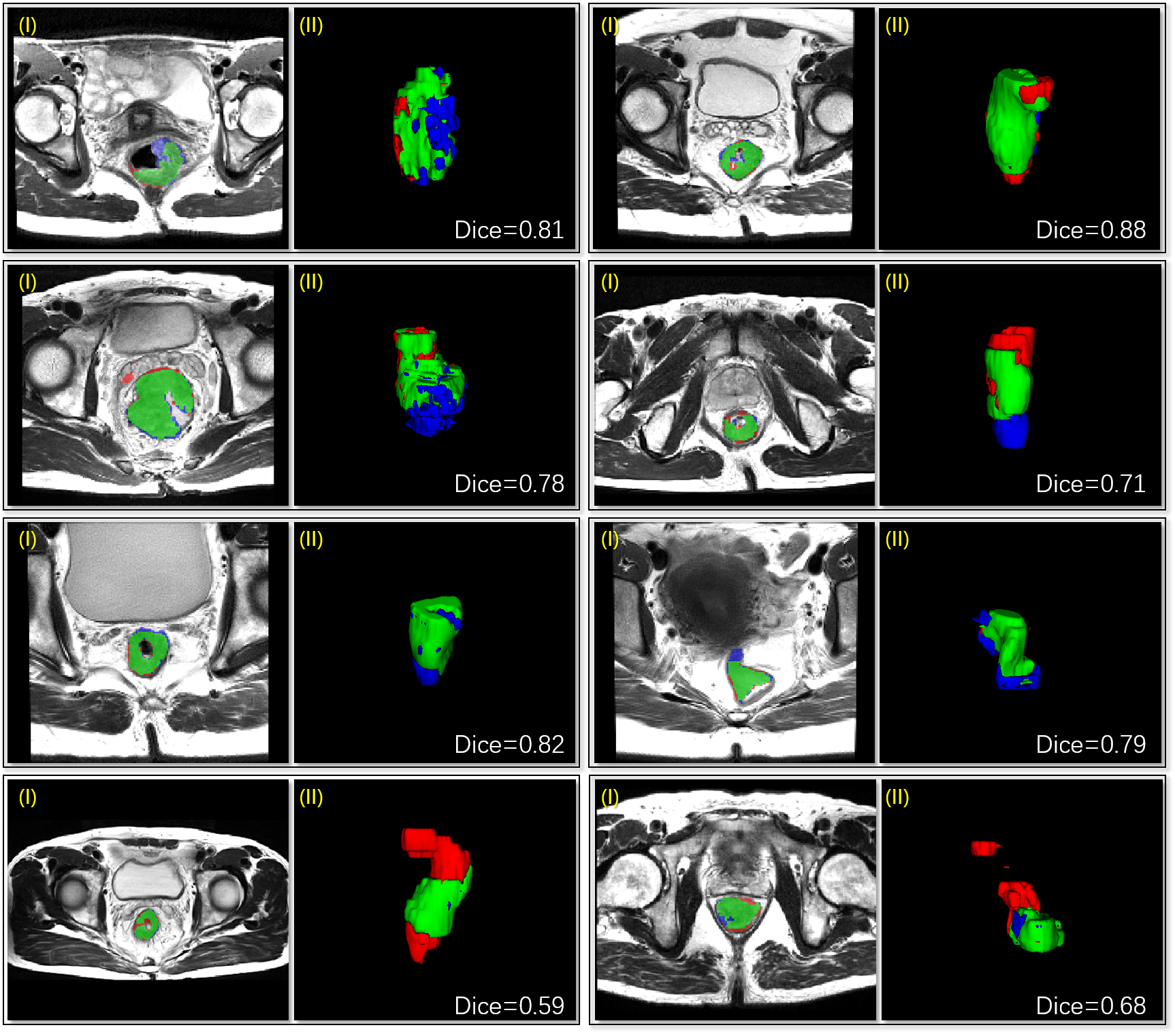}
	
	\caption{{\color{black} (I) selected 2D slices (II) 3d segmentation masks. Green indicates true positives; Red indicates false positives; Blue indicates false negatives.} }
	\label{fig:Results2}
\end{figure*}
\begin{table*}
	
	\caption{A{\upshape blation studies and comparisons. }}
	\centering
	\label{tab:compare}
	\resizebox{0.95\textwidth}{!}{
	\begin{tabular}{lcccccc}
		\toprule[1pt]
		Method & DSC[\%] & Recall[\%] & ASD[mm] & $Avg N_{RoIs}$ & Loc/Seg GPU Time[s] & Loc/Seg CPU Time[s]\\
		\midrule
		{\color{black}\textbf{3D RU-Net+MHL+Ensemble}} & \textbf{75.5$\pm$10.7} &\textbf{77.8$\pm$14.8}& \textbf{2.45$\pm$3.26} & - & 0.61&38.10\\
        {\color{black}3D RU-Net-RF112+MHL (HighRes)} & 74.2$\pm$10.6 &78.6$\pm$13.9 & 3.02$\pm$3.55 & 1.8 &$0.22/0.01$&$15.30/0.79$\\
		{\color{black}3D RU-Net-RF88+MHL (HighRes)} &  73.7$\pm$10.5 &75.4$\pm$14.8 & 2.83$\pm$3.57 & 2.2 & $0.17/0.01$&$10.44/0.45$\\
		{\color{black}3D RU-Net-RF64+MHL (HighRes)} & 72.7$\pm$12.5 &76.2$\pm$17.2 & 2.62$\pm$3.05 & 2.7 & $0.15/0.01$&$8.98/0.34$\\
		
		{\color{black}3D FCN+3D U-Net+HL+Ensemble} & 73.4$\pm$11.6&78.5$\pm$15.3 & 3.10$\pm$3.71 & - & 0.72 & 44.45\\

		{\color{black}3D FCN+3D U-Net-RF122+HL (HighRes)} & 72.0$\pm$12.1&78.2$\pm$15.5 & 3.41$\pm$4.14 & 2.2 & $0.22/0.03$ & $15.30/1.77$\\

		{\color{black}3D FCN+3D U-Net-RF88+HL (HighRes)} & 71.6$\pm$12.5&75.9$\pm$16.6 & 3.42$\pm$3.73 & 2.6 & $0.17/0.02$ & $10.44/1.12$\\
		
		{\color{black}3D FCN+3D U-Net-RF64+HL (HighRes)} & 71.7$\pm$11.9&79.1$\pm$14.9 & 3.56$\pm$4.07 & 3.1 & $0.15/0.02$ & $8.98/0.96$\\
		3D U-Net-RF64+HL+Ensemble\cite{Huang2018HL} & 72.1$\pm$13.9&72.2$\pm$17.2 & 3.83$\pm$4.95 & - & 18.11 & 616.70 \\
		3D U-Net-RF64+HL\cite{Huang2018HL}~(LowRes) & 69.9$\pm$12.5&70.2$\pm$14.9  & 3.90$\pm$4.43& - & 2.25 & 88.90\\
		3D U-Net-RF64+HL\cite{Huang2018HL}~(MidRes) & 70.0$\pm$14.5&72.1$\pm$17.3 & 5.48$\pm$7.06 & - & 5.60 & 180.62\\
		3D U-Net-RF64+HL\cite{Huang2018HL}~(HighRes) & 67.7$\pm$18.4&69.2$\pm$21.3 & 10.24$\pm$14.59 & - & 10.26 & 346.72\\
		{\color{black}3D U-Net-RF112+HL\cite{Huang2018HL}~(HighRes)} & 67.8$\pm$13.8&70.5$\pm$14.8 & 12.44$\pm$13.70 & - & 14.62 & 827.57\\
		{\color{black}3D U-Net-RF88+HL\cite{Huang2018HL}~(HighRes)} & 66.9$\pm$17.4&71.8$\pm$15.7 & 14.16$\pm$11.54 & -& 12.47 & 358.88\\
		3D 
		3D U-Net-RF64+DL+Ensemble\cite{milletari2016v} & 69.9$\pm$13.7&72.4$\pm$18.0 & 4.18$\pm$5.89 & - &18.11 & 616.70\\
		3D U-Net-RF64+DL\cite{milletari2016v}~(LowRes) & 68.5$\pm$13.8&68.5$\pm$19.5 & 4.19$\pm$5.75 & - & 2.25 & 88.90\\
		3D U-Net-RF64+DL\cite{milletari2016v}~(MidRes) & 67.3$\pm$15.3&70.2$\pm$17.7 & 5.70$\pm$7.31 & - & 5.60 & 180.62\\
		3D U-Net-RF64+DL\cite{milletari2016v}~(HighRes) & 66.0$\pm$18.2&70.9$\pm$22.0 & 10.32$\pm$12.11 & - & 10.26 &346.40\\
		3D U-Net-RF64\cite{cciccek20163d}~(HighRes)  & 61.7$\pm$19.2&57.3$\pm$23.9& 4.26$\pm$4.35& -& 10.26 & 346.40\\
		\midrule
		{\color{black}2D U-Net+3D U-Net+Ensemble\cite{Balagopal2018Fully}} & 72.0$\pm$13.6&76.1$\pm$18.3 & 3.86$\pm$5.46 & - & 1.021 & 79.70\\
		{\color{black}2D U-Net+3D U-Net\cite{Balagopal2018Fully}~(LowRes)} & 70.2$\pm$12.4&74.5$\pm$15.8 & 4.11$\pm$5.03 & 5.3 & $0.15/0.02$ & $7.68/0.53$\\
		{\color{black}2D U-Net+3D U-Net\cite{Balagopal2018Fully}~(MidRes)} & 69.1$\pm$17.7&73.7$\pm$21.0 & 6.05$\pm$9.53 & 6.1 & $0.18/0.02$&$16.95/0.87$\\
		{\color{black}2D U-Net+3D U-Net\cite{Balagopal2018Fully}~(HighRes)} & 69.4$\pm$14.1&76.2$\pm$18.2 & 6.23$\pm$8.77 & 7.1 &  $0.25/0.03$ & $38.29/1.22$\\
		{\color{black}2D kU-Net+BDC-LSTM\cite{chen2016combining}~(HighRes)} & 69.3$\pm$13.1&79.1$\pm$16.7 & 7.81$\pm$6.88 & - & 0.51 & 39.22\\
		{\color{black}Super-Voxel Clustering\cite{irving2014automated}~(HighRes)} & 62.6$\pm$14.9&60.2$\pm$18.2 & 6.54$\pm$5.96 & - & - & 15.13\\
		3D Mask R-CNN\cite{He2017Mask}~(HighRes) & 56.4$\pm$19.0&58.5$\pm$25.6 & 7.93$\pm$10.33 & - & 0.55 & 35.88\\
		{\color{black}3D Mask R-CNN\cite{He2017Mask}~(MidRes)} & 54.6$\pm$17.3&61.0$\pm$24.5 & 9.05$\pm$8.53 & - & 0.32 & 18.07\\
		{\color{black}3D Mask R-CNN\cite{He2017Mask}~(LowRes)} & 52.0$\pm$16.8&55.0$\pm$24.1 & 7.02$\pm$7.24 & - & 0.24 & 11.61\\
		\bottomrule[1pt]
		
	\end{tabular}
  	}
\end{table*}

\begin{table*}
	
	\caption{GPU{\upshape\ memory footprint tracking given an input volume of size $40\times180\times320$ and RoI size of $24\times96\times96$}}
	\centering
	\label{tab:dram}
	\resizebox{\textwidth}{!}{
    	\begin{tabular}{c|ccc|c}
    		\toprule[1pt]
    		Part Name &Layer Name & Size & GPU Memory Footprint & Part GPU Memory Footprint\\  
    		\midrule
    		\multirow{6}{2cm}{Encoder}&ResBlock1&$9\ nodes\times40\times180\times320\times48\ channels$& 3796.88 MBytes&\multirow{6}{2cm}{6302.047MBytes}\\
    		&MaxPooling1&$1\ node\times40\times90\times160\times48\ channels$&105.47 MBytes&\\
    		&ResBlock2&$9\ nodes\times40\times90\times160\times96\ channels$&1898.45 MBytes&\\
    		&MaxPooling2&$1\ node\times20\times45\times80\times96\ channels$&26.37 MBytes&\\
            &ResBlock3&$9\ nodes\times20\times45\times80\times192\ channels$&474.60 MBytes&\\
    		&Locator~(sigmoid)&$1\ node\times20\times45\times80\times1\ channel$&0.27MBytes&\\
    		\midrule
    		\multirow{3}{2cm}{RoI Tensor Pyramid} & RoI Tensor1 & $1\ node\times24\times96\times96\times48\ channels$ & 40.50 MBytes &\multirow{3}{2cm}{65.82 MBytes}\\
    		&RoI Tensor2 & $1\ node\times24\times48\times48\times96\ channels$ & 20.25 MBytes &\\
    		&RoI Tensor3 & $1\ node\times12\times24\times24\times192\ channels$ & 5.06 MBytes &\\

    		\midrule
    		\multirow{8}{2cm}{Local Region Decoder}&UpConv1&$1\  node\times24\times48\times48\times96\ channels$ & 20.25 MBytes&\multirow{8}{2cm}{669.93 MBytes}\\
    		&Add1 & $1\ node\times24\times48\times48\times96\ channels$ & 20.25 MBytes &\\
    		&ResBlock4 & $9\ nodes\times24\times48\times48\times96\ channels$ & 182.25 MBytes &\\
    		&UpConv2 & $1\ node\times24\times96\times96\times48\ channels$ & 40.50 MBytes &\\
    		&Add2 & $1\ node\times24\times96\times96\times48\ channels$ & 40.50 MBytes &\\
    		&ResBlock5 & $9\ nodes\times24\times96\times96\times48\ channels$ & 364.50 MBytes &\\
    		&SegHead1~(sigmoid) & $1\ node\times24\times96\times96\times1\ channels$ & 0.84 MBytes &\\
    		&SegHead2~(sigmoid) & $1\ node\times24\times96\times96\times1\ channels$ & 0.84 MBytes &\\		

    		\midrule
    		\multirow{8}{2cm}{{\color{black}Standard Decoder}}&UpConv1&$1\  node\times40\times90\times160\times96\ channels$ & 210.94 MBytes&\multirow{8}{2cm}{6978.55 MBytes}\\
    		&Add1 & $1\ node\times40\times90\times160\times96\ channels$ & 210.94 MBytes &\\
    		&ResBlock4 & $9\ nodes\times40\times90\times160\times96\ channels$ & 1898.45 MBytes &\\
    		&UpConv2 & $1\ node\times40\times180\times320\times48\ channels$ & 421.88 MBytes &\\
    		&Add2 & $1\ node\times40\times180\times320\times48\ channels$ & 421.88 MBytes &\\
    		&ResBlock5 & $9\ nodes\times40\times180\times320\times48\ channels$ & 3796.88 MBytes &\\
    		&SegHead1~(sigmoid) & $1\ node\times40\times180\times320\times1\ channels$ & 8.79 MBytes &\\
    		&SegHead2~(sigmoid) & $1\ node\times40\times180\times320\times1\ channels$ & 8.79 MBytes &\\
    		\bottomrule[1pt]
	    \end{tabular}
        }
\end{table*}
\subsection{Results}

For evaluation, four-fold cross-validation was conducted on 64 scans {\color{black}and their mean scores are reported in TABLE. \ref{tab:compare}.} Comparison of predicted masks between different methods is illustrated in Fig. \ref{fig:Results1}; Eight volume predictions are illustrated in Fig. \ref{fig:Results2}. 

{\color{black}\subsubsection{Ablation Studies} Firstly, we conduct a full ablation study to evaluate the contribution of each proposed component, listed in the upper section of TABLE. \ref{tab:compare}. 

Compared to the part based 3D U-Net~\cite{cciccek20163d} built with the ResBlocks described in \ref{Design}, 3D U-Net-RF64+DL~\cite{milletari2016v} and 3D U-Net-RF64+HL~\cite{Huang2018HL} using Dice loss and hybrid loss improved the performance by alleviating the class imbalance problem. Specifically, we acquired $(d, h, w)=(24,96,96)$ patches at a stride of 50\% window overlapping for training and predicting and found that despite that details are sacrificed, down-sampling, $i.e.$ using MidRes and LowRes image sets, significantly boosts the performance due to enlarged physical receptive fields. As a step further, can we enlarge the receptive field defined by the network's convolution kernels rather than down-sampling the images for better performance without sacrificing details? Following the criterion stated in \ref{ensemble}, we tuned dilation rates of 3D U-Net+HL to form 3D U-Net-RF64+HL, 3D U-Net-RF88+HL and 3D U-Net-RF112+HL, with receptive fields of $26 \times 64 \times 64$, $26 \times 88 \times 88$ and $26 \times 122 \times 122$, respectively. The experimental results demonstrate that enlarging receptive field without enlarging patches does not take the performance to the level of down-sampling based methods. These results highlight that the input volume size hindered the receptive field from taking advantage of wide range contexts. 

Apparently, the pipeline can be accelerated by employing either of a non-joint or a joint detetion-segmentation framework, and most of the false positives can be eliminated as well. To further emphasize the merit of whole volume joint training and cross-module feature sharing enabled by the proposed method named as 3D RU-Net+MHL, a detection-segmentation cascaded model without these properties, namely 3D FCN+3D U-Net, is designed and evaluated. The cascaded model consists of a standalone Global Image Encoder for RoI detection and a full 3D U-Net replacing the Local Region Decoder for in-region segmentation, trained towards Dice loss and Hybrid loss, respectively. We also tuned the 3D U-Net's receptive field as $26 \times 64 \times 64$, $26 \times 88 \times 88$ and $26 \times 122 \times 122$, and did not notice a significant performance difference~($<0.4\%$). On the other hand, the proposed method, however, enjoyed a higher Dice score~(from 72.7\% to 74.2\%) by enlarging the receptive field, which is a significant performance boost considering that no extra parameter or module is included. Another merit of feature sharing lies on the observation that a joint trained model's Locator module and Local Region Decoder module share a highly consistent behavior pattern except for detail richness. On the contrary, the cascaded model suffers from more in-region false positives due to the non-joint training scheme and the patch size limit of the receptive field, therefore produced higher recall rates along with more false positives (larger $Avg N_{RoIs}$) and scored lower mean Dice and larger mean ASD. 

Nevertheless, the proposed method is significantly faster than part-based methods, which is almost impossible for CPU-only deployment, while our method costs 15 seconds to score 74.2\% mean Dice and less than 40 seconds to achieve 75.5\% mean Dice, providing faster and more accurate predictions. Furthermore, the proposed Local Region Decoder is 2X faster compared to the cascaded model's segmentation branch, which makes significant difference when an image carries multiple detected RoIs. 

Aforementioned performance gains and speedups are enabled by improved memory efficiency: compared to vanilla 3D U-Net, the largest volume size trainable using a device with 12GB GPU memory is increased from $48\times 168 \times 168$ to $48\times 288 \times 288$, under aforementioned fixed parameter setting.} 

{\color{black}\subsubsection{Cross-Methodology Comparison} Next, we conducted cross-methodology evaluation by comparing the proposed method to other third-party methodologies. 

Firstly, 2D U-Net+3D U-Net proposed by \cite{Balagopal2018Fully} is another version of model cascading. Compared to 3D FCN+3D U-Net and 3D RU-Net, a 2D U-Net serving as an RoI locator produces significantly more false positive candidates (larger $Avg N_{RoIs}$) with larger length along the Z-axis, which degrades the performance and is more time costly.

Next, a 2D U-Net+BDC-LSTM~\cite{chen2016combining} is evaluated, whose kU-Net is employed for intra-slice feature extraction and a bidirectional convolutional LSTM is used to explore intra-slice features. Since patch size no longer limits the effective receptive field, we evaluated this method only using the HighRes dataset with a large designed receptive field as is proposed in \cite{chen2016combining}. It scored similarly compared to a 3D U-Net+HL, highlighting the effectiveness of intra-slice LTSMs. However, it only partially resolved the problem and got higher recall rate along with larger ASD since that full 3D context utilization is still limited and more false positives are produced along the Z axis, and its execution time is significantly longer compared to the proposed method. 

Additionally, a 3D-FPN based Mask R-CNN is evaluated. As the scores and figures illustrates, the limitation of 3D Mask R-CNN is two-fold: bad-shaped bounding boxes' cutting off some parts of the objects, and low-resolution masks generated by the coarse-resolution feature maps of the FPN backbone. 

Finally, we also set a super-voxel clustering based~\cite{irving2014automated} method as the baseline. Without the merit of discriminative 3D deep features, super-voxels are inevitably over-segmented or under-segmented. In our experiments, one of the 64 targets went completely missing and significantly lowered the Dice score, while some wrong super-voxels were chosen as the output mask.}

\section{Discussion}

In this paper, we proposed a method to inherit easy-to-train and detail-preserving merits of volume-to-volume 3D FCNs while acquiring fast RoI localization, target completeness and whole volume global understanding of a joint detection-segmentation framework enabled by its large receptive field shared across different tasks. We combined a whole volume RoI localization model named as Global Image Encoder and in-region segmentation model named as Local Region Decoder as a joint model named 3D RoI-aware U-Net~(3D RU-Net). As the result, we could segment colorectal tumors accurately and fast. 

We notice a recent trend that researches seek to segment medical objects via detection. But most of these works employ independent modules for different tasks, leaving the benefit of fast feature reusing and wide range context utilization of large receptive field not fully enjoyed. As a refinement, the proposed method utilize the pre-extracted globally encoded features for in-region segmentation, providing better understanding of the whole image in the segmentation branch to discriminate background from false positives, and further saved over 50\% computing resource for each in-region segmentation, which significantly accelerated the workflow in circumstances where multiple targets are detected.

Compared to successful and general Mask R-CNN for natural object instance segmentation, the advantage of the proposed framework over Mask R-CNN mainly lies on {\color{black}its full utilization of voxel-wise labels for target detection, and the lossless segmentation process similar to volume-to-volume 3D FCNs that fully restores the targets' dimension. Hindered by additional aspect ratios and number-limited training samples, it's sub-optimal to degrade voxel-wise labels to target-wise labels, hence bad-shaped bounding boxes are frequently predicted and cut off parts of the targets. Plus, the detail lossing in-region segmentation scheme of Mask R-CNN also produced inferior detail richness compared to end-to-end FCNs. }

Finally, it's significant to point out that the speed and performance gains are enabled by the memory efficiency of the proposed method that eliminates the need of conventional 3D U-Net for sliding-stitching workflow and enables one-step whole volume inference. Here we track the memory footprint to evaluate the memory efficiency of the proposed method in the environment where in-place computing is deactivated thus a ResBlock has nine tensor nodes. Given a typical T2 volume of 3D pelvic image of size $40\times320\times320$, by body cropping, the size typically drops to $40\times180\times320$. With this volume as input, the GPU memory footprint details are listed in TABLE. \ref{tab:dram}. By constructing the Local Region Decoder, a GPU can assign 90\% of its GPU memory to the encoder to process larger volumes and spend only 10\% GPU memory on the segmentation stage, while conventional encoder-decoder networks spend ~50\% GPU memory on each path, {\color{black}as is hypothetically computed in the Standard Decoder section of TABLE. \ref{tab:dram}}. Therefore, the applicable volume size is dramatically enlarged. In addition, while model ensemble strategy is often considered to be computationally expensive, based on the proposed method, we can have the performance gain at a promisingly acceptable cost.

Although our method achieved competitive results, there are some limitations. 
Firstly, as is illustrated in Fig. \ref{fig:Results2}, the model is often confused about which slice to start or end, thus this significantly affects the score. {\color{black}As is illustrated in TABLE. \ref{tab:compare}, all competing methods including applying a bidirectional convolutional LSTM~\cite{chen2016combining} did not thoroughly tackle this issue.} As an explanation, this difficulty is data-related and decision about starting and ending slice index can be observer-dependent due to weak contrast in the border of cancerous tissues and low resolution along Z axis.  {\color{black}Secondly, for this specific task without the need of discriminating different instances of tumors, we did not include instance separation capability in our design. However, it can be addressed since that the RoI Locator is a template that any encoder-only detection method is applicable, yet it can still be beneficial to fully utilize voxel-wise labels for bounding box refinement.}


\section{Conclusion}

In this paper, we proposed a joint RoI localization-segmentation-based framework for fully automatic one-step whole volume colorectal cancer segmentation referred to as 3D RoI-aware U-Net~(3D RU-Net). We emphasized the importance and effectiveness of integrating RoI localization and in-region segmentation fed with globally encoded features to perform fast and accurate whole volume segmentation. The proposed method enables the merit of enlarging receptive fields originally limited by GPU memory capacity and ensemble models with different receptive field settings. A Dice-formulated multi-task hybrid loss function is present to smoothen the training process. Experimental results demonstrated impressive superiority in terms of accuracy and speed over competing methods. In principle, the proposed framework is scalable enough to be adopted to other medical image segmentation tasks.

\bibliographystyle{IEEEbib}
\bibliography{refs}



\end{document}